\documentclass{bmvc2k}

\title{Feature Binding with Category-Dependant MixUp for Semantic Segmentation and Adversarial Robustness}

\addauthor{Md Amirul Islam}{www.cs.ryerson.ca/~amirul}{1,2}
\addauthor{Matthew Kowal}{mkowal2.github.io}{1}
\addauthor{Konstantinos G. Derpanis}{www.cs.ryerson.ca/~kosta}{1,2,3}
\addauthor{Neil D. B. Bruce}{www.cs.ryerson.ca/~bruce}{1,2,4}

\addinstitution{
 Ryerson University\\
 Toronto, Canada
}
\addinstitution{
 Vector Institute\\
 Toronto, Canada
}
\addinstitution{
 Samsung AI Centre\\
 Toronto, Canada
}
\addinstitution{
 University of Guelph\\
 Guelph, Canada
}
\runninghead{Islam, Kowal, Derpanis, Bruce}{Category Dependent MixUp}
\usepackage{graphicx}
\usepackage{amsmath}
\usepackage{amssymb}
\usepackage{booktabs}
\usepackage{algpseudocode}
\usepackage{algorithm}
\usepackage{mathtools}
\usepackage{color}
\usepackage{multirow,graphicx,paralist}
\usepackage{ colortbl}
\usepackage{ tabularx}
\usepackage{pgfplots}
\usepackage{enumitem}
\pgfplotsset{compat=1.3}
\definecolor{Gray}{gray}{0.85}
\definecolor{LightCyan}{rgb}{0.88,1,1}
\definecolor{antiquefuchsia}{rgb}{0.57, 0.36, 0.51}
\definecolor{bleudefrance}{rgb}{0.19, 0.55, 0.91}
\usepackage[export]{adjustbox}
\usepackage[font=small,labelfont=bf]{caption}
\definecolor{maroon}{cmyk}{0,0.87,0.68,0.32}

\usepackage{upquote}

\begin{document}

\maketitle

\begin{abstract}
In this paper, we present a strategy for training convolutional neural networks to effectively resolve interference arising from competing hypotheses relating to inter-categorical information throughout the network. The premise is based on the notion of feature binding, which is defined as the process by which activation's spread across space and layers in the network are successfully integrated to arrive at a correct inference decision. In our work, this is accomplished for the task of dense image labelling by blending images based on their class labels, and then training a \textit{feature binding} network, which simultaneously segments and separates the blended images. Subsequent feature denoising to suppress noisy activations reveals additional desirable properties and high degrees of successful predictions. Through this process, we reveal a general mechanism, distinct from any prior methods, for boosting the performance of the base segmentation network while simultaneously increasing robustness to adversarial attacks.


\end{abstract}
\begin{figure}[h]
\vspace{-0.3cm}
	\begin{center}
		\includegraphics[width=0.98\textwidth]{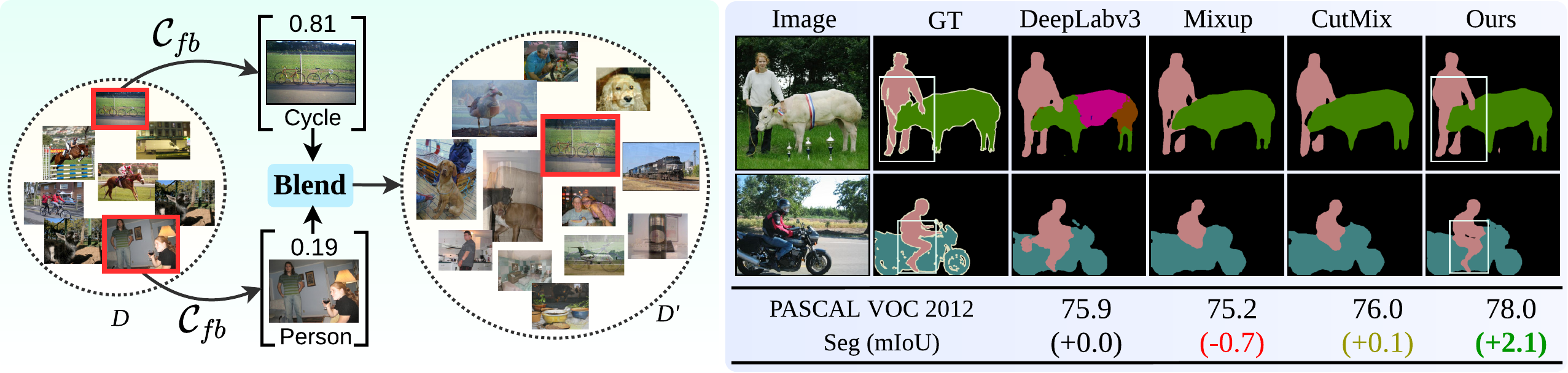}
		\vspace{0.3cm}
		\caption{Left: Overview of our category-specific ($\mathcal{C}_{fb}$) image blending to create a new source dataset ($D'$). A segmentation network is trained with $D'$ to simultaneously separate and segment both source images. Right: Results of our \textit{feature binding} method, Mixup~\cite{zhang2017mixup}, and CutMix~\cite{yun2019cutmix} on the PASCAL VOC 2012~\cite{everingham2015pascal} segmentation task. Note that, our method significantly improves the performance.}
		\label{fig:intro}
	\end{center}
	\vspace{-0.6cm}
\end{figure}

\vspace{-0.2cm}
\section{Introduction}\label{sec:intro}
The advent of Deep Neural Networks (DNNs) has seen overwhelming improvement in dense image labeling tasks~\cite{long15_cvpr,noh15_iccv, badrinarayanan15_arxiv,ghiasi2016laplacian,zhao2017pyramid,Islam_2017_CVPR,islam2017label,chen2018deeplab,islam2018gated,cvpr18_rank,he2017mask,islam2018semantics, li2016iterative,karim2019recurrent,karim2020distributed}, however, for some common benchmarks~\cite{everingham2015pascal} the rate of improvement has slowed down. While one might assume that barriers to further improvement require changes at the architectural level, it has also been borne out that pre-training across a variety of datasets~\cite{russakovsky2015imagenet,lin2014microsoft} can improve performance exceeding improvements seen from changing the model architecture. However, there are challenging scenarios for which DNNs have difficulty on regardless of pre-training or architectural changes, such as highly occluded scenes, or objects appearing out of their normal context~\cite{singh2020don}. It is not clear though, for dense image labeling tasks, how to resolve these specific scenarios for more robust prediction quality on a per-pixel level.

A question that naturally follows from this line of reasoning is: How can the number of locally challenging cases be increased, or the problem made more difficult in general? In this paper, we address this problem using a principled approach to improve performance and that also implies a more general form of robustness. As inspiration, we look to a paradigm discussed often in the realm of human vision: the binding problem~\cite{treisman1998feature,shipp2009feature}. The crux of this problem is that given a complex decomposition of an image into features that represent different concepts, or different parts of the image, how does one proceed to successfully relate activations corresponding to common sources in the input image to label a \emph{whole} from its parts, or separate objects. Motivated by the binding problem, a successful solution in the computer vision domain should rely on both determining correspondences in activations among features that represent disparate concepts, and also to associate activations tied to related features that are subject to spatial separation in the image. To address similar issues for the image classification task, recent studies~\cite{tokozume2018between,zhang2017mixup,yun2019cutmix} have considered mixing two image examples with constraints on the distribution of features. However, these methods suffer from biases in the dataset used, as they have no strategy when deciding on which images to mix which is crucial for the dense labeling problem. Additionally, these strategies do not adequately separate information from different sources in the image as they only require the network to make a single (classification) prediction during training.

In our work, the means of solving the feature binding problem takes a direct form, which involves training networks on a specially designed dataset of mixed images to simultaneously address problems of dense image labeling~\cite{long15_cvpr,chen15_iclr,noh15_iccv}, and blind source separation~\cite{georgiev2005sparse,huang2015joint}. Humans show a surprising level of capability in interpreting a superposition (e.g., average) of two images, both interpreting the contents of each scene and determining the membership of local patterns within a given scene. The underlying premise of this work involves producing networks capable of simultaneously performing dense image labeling for pairs of images while also separating labels according to the source images. If one selects pairs on the basis of a weighted average (see Fig.~\ref{fig:intro} (left)), this allows treatment of the corresponding dense image labeling problem in the absence of source separation by extension. This process supports several objectives: (i) it significantly increases the number of occurrences that are locally ambiguous that need to be resolved to produce a correct categorical assignment, (ii) it forces broader spatial context to be considered in making categorical assignments, and (iii) it stands to create more powerful networks for standard dense labeling tasks and dealing with adversarial perturbations by forcing explicit requirements on how the network uses the input. The end goal of our procedure is to improve overall performance as well as increase the prediction quality on complex images (see Fig.~\ref{fig:intro} (right)), heavily occluded scenes, and also invoke robustness to challenging adversarial inputs. 
Our main contributions are as follows:
\vspace{-0.1cm}
\begin{itemize}
  \setlength{\itemsep}{1.3pt}
    \item To the best of our knowledge we present the first work which applies image blending to the dense labeling task. To this end, we propose a novel training pipeline which simultaneously solves the problems of dense labeling and blind source separation. 
    \item We further introduce a new \textit{categorical clustering} strategy which exploits semantic knowledge of the dataset to mix input images based on their class distributions.
    \item We show, through extensive quantitative and qualitative experiments, that our pipeline outperforms recent image blending methods~\cite{zhang2017mixup,yun2019cutmix} on the PASCAL VOC 2012 dataset~\cite{everingham2015pascal}, while simultaneously improving robustness to adversarial attacks.
\end{itemize}

\section{Related Work}\label{sec:related}

More closely related to the feature binding concept, contributions~\cite{yun2019cutmix,zhang2017mixup,tokozume2018between,inoue2018data,cubuk2019autoaugment,french2019semi,harris2020fmix,chou2020remix} on data augmentation based techniques share a similar idea of mixing two randomly selected samples to create new training data for the image classification or localization task. BC learning~\cite{tokozume2018between} showed that randomly mixing training samples can lead to better separation between categories based on the feature distribution. Mixup~\cite{zhang2017mixup} shares a similar idea of training a network by mixing the data that regularizes the network and increases the robustness against adversarial examples, whereas CutMix~\cite{yun2019cutmix} proposed to overlay a cropped area of an input image to another. Our proposed \textit{feature binding} approach differs from the above existing works in that: (i) the network performs simultaneous dense prediction and blind source separation to achieve superior dense labeling and adversarial robustness whereas other techniques are focused mainly on image classification or object localization, (ii) previous methods either mix labels as the ground truth or use the label from only one sample, while we use both ground truth labels independently, and (iii) samples are chosen randomly for Mixup~\cite{zhang2017mixup} and CutMix~\cite{yun2019cutmix} while we use an intuitive strategy (categorical clustering, Sec.~\ref{sec:fbt}).
\section{Proposed Method}\label{sec:approach}
In the broader context of investigating approaches motivated by the feature binding problem, we propose a novel framework capable of solving the dense labeling problem. Our proposed framework consists of three key steps: (i) we first apply a technique on the training dataset that generates a new set of source images (Sec.~\ref{sec:fbt}), (ii) we train a convolutional neural network (CNN) using the generated data that produces dense predictions (Sec.~\ref{sec:FBNet}), and (iii) we denoise the learned features from the feature binding process by fine-tuning on the standard data (Sec.~\ref{sec:denoise}). 
\subsection{Category-Dependent Image Blending}\label{sec:fbt}
Recent works~\cite{zhang2017mixup,yun2019cutmix,tokozume2018between,inoue2018data,cubuk2019autoaugment} simply mix two randomly selected samples to create new training data for classification or localization task. Exploring a similar direction, we are interested in solving dense prediction in a way that provides separation based on mixed source images. We augment the PASCAL VOC 2012~\cite{everingham2015pascal} training dataset via a novel data processing stage to generate a new training set in a form that accounts for source separation and dense prediction. The traditional way~\cite{zhang2017mixup,tokozume2018between,inoue2018data} of combining two images is by weighted average which implies that the contents of both scenes appear with varying contrast. Randomly combining two source images to achieve the desired objective is a more significant challenge than one might expect in the context of dense prediction. One challenge is the categorical bias of the dataset (e.g., mostly the \textit{person} images will be combined with all other categories since \textit{person} is the most common category in PASCAL VOC 2012) across the newly generated training set. Previous methods~\cite{zhang2017mixup,tokozume2018between,inoue2018data}, randomly select images to combine, results in a new data distribution which inherit similar biases as the original dataset. To overcome these limitations, we propose a technique denoted as \textit{categorical clustering}, $\mathcal{C}_{f_b}$, which combines images based on a uniform distribution across categories. Thorough experimentation with our proposed mixing strategy show improvements in the network's ability to separate competing categorical features and can generalize these improvements to various challenging scenarios, such as segmenting out-of-context objects or highly occluded scenes.

\noindent \textbf{Categorical Clustering:} We first generate 20 different clusters of images where each cluster contains images of a certain category from VOC 2012. For each training sample in a cluster, we linearly combine it with a random sample from each of the 19 other clusters. For example, given a training sample $\mathcal{I}^1_{s}$ from the \textit{person} cluster we randomly choose a sample $\mathcal{I}^2_{s}$ from another categorical cluster and combine them to obtain a new sample, $\mathcal{I}_{fb}$: 
\vspace{-0.1cm}
\begin{gather}\label{eq:combine}
   \mathcal{I}_{fb} = \delta \ast \mathcal{I}^1_{s} + (1-\delta) \ast \mathcal{I}^2_{s} ,
   \vspace{-0.4cm}
\end{gather}
where $\delta$ denotes the randomly chosen weight that is applied to each image. We assign the weight such that the source image ($\mathcal{I}^1_{s}$) has more weight compared to the random one ($\mathcal{I}^2_{s}$). In our experiments, we sample $\delta$ uniformly from a range of $[0.7-1]$ for each image pair. We also change the range of $\delta$ values and report results in Table~\ref{tab:fb_quan} (a). Note that, for one sample in \textit{person} cluster we generate 19 new samples. We continue to generate feature binding samples for the other remaining images in the person cluster and perform the same operation for images in other clusters. While there may exist alternatives~\cite{zhang2017mixup,devries2017improved,yun2019cutmix} for combining pairs of images to generate a training set suitable for source separation training, our intuitive method is simple to implement and achieves strong performance on a variety of metrics (see Section~\ref{sec:exp}). Exploring further methods to combine and augment the training set is an interesting and nuanced problem to be studied further in dense image labelling.

\subsection{Feature Binding Network}\label{sec:FBNet}
\begin{figure}
	\begin{center}
		\includegraphics[width=0.98\textwidth]{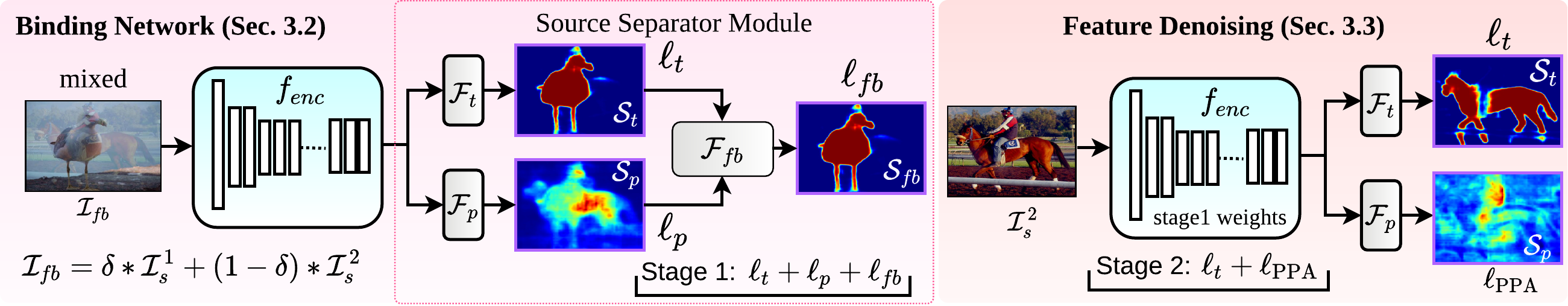}
		\vspace{0.3cm}
		\caption{An illustration of \textit{feature binding} process. At the data end, categorical collisions are created with a dominant ($\mathcal{I}_{s}^{1}$) and phantom ($\mathcal{I}_{s}^{2}$) image. \textbf{Stage 1:} The network is trained on mixed data ($\mathcal{I}_{fb}$) to perform simultaneous dense labeling and source separation. We use the labels of both source images as the targets for two separate output channels. \textbf{Stage 2:} Fine tuning on standard data to further promote desirable properties along the two dimensions of base performance and robustness to perturbations. In this stage, the phantom activation of the second channel is suppressed. Confidence maps are plotted with the `Jet' colormap, where red and blue indicates higher and lower confidence, respectively.}
		\label{fig:architecture}
	\end{center}
	\vspace{-0.5cm}
\end{figure}

In this section, we present a fully convolutional feature binding network in the context of dense prediction. Fig.~\ref{fig:architecture} illustrates the overall pipeline of our proposed method. 
Fig.~\ref{fig:architecture} (left) reveals two key components of the binding network including a \textit{fully convolutional network} encoder and \textit{source separator module} (SSM). Given a mixed image $\mathcal{I}_{fb}{\in \mathbb{R}^{h\times w \times c}}$, we adopt a DeepLabv3~\cite{chen2017rethinking} ($f_{enc}$) to produce a sequence of bottom-up feature maps. The SSM consists of two separate branches: (i) \textit{dominant} $\mathcal{F}_t(.)$, and (ii) \textit{phantom}, $\mathcal{F}_{p}(.)$. Each branch takes the spatial feature map, $\hat{f}_{b}^i$, produced at the last block, \texttt{res5c}, of $f_{enc}$ as input and produces a dense prediction for the dominant, $\mathcal{S}_{t}$, and the phantom, $\mathcal{S}_p$, image. Next, we append a \textit{feature binding head} (FBH) to generate a final dense prediction of categories for the dominant image. The FBH, $\mathcal{F}_{fb}$, simply concatenates the outputs of source and phantom branches followed by two $1\times 1$ convolution layers with non-linearities (ReLU) to obtain the final dense prediction map, $\mathcal{S}_{fb}$. The intuition behind the FBH is that the phantom branch may produce activations that are correlated with the dominant image, and thus the FBH allows the network to further correct any incorrectly separated features with an additional signal to learn from. Given a mixed image, $\mathcal{I}_{fb}$, the operations can be expressed as:
\begin{gather}
\hat{f}_{b}^i = f_{\textit{enc}}(\mathcal{I}_{fb}), \hspace{0.2cm} \underbrace{\mathcal{S}_t=\mathcal{F}_t(\hat{f}_{b}^i)}_\texttt{dominant}, \hspace{0.2cm}  \underbrace{\mathcal{S}_{p}=\mathcal{F}_{p}(\hat{f}_{b}^i)}_\texttt{phantom}, \hspace{0.2cm} 
\underbrace{\mathcal{S}_{fb}=\mathcal{F}_{fb}(\mathcal{S}_t, \mathcal{S}_{p})}_\texttt{binding}.
\end{gather}
\noindent \textbf{Training the Feature Binding Network.}
The feature binding network produces two dominant predictions, $\mathcal{S}_{fb}$ and $\mathcal{S}_t$, including a phantom prediction, $\mathcal{S}_p$; however, we are principally interested in the final dominant prediction, $\mathcal{S}_{fb}$. In more specific terms, let $\mathcal{I}_{fb}\in{\rm I\!R}^{h\times w \times 3} $ be a training image associated with ground-truth maps ($\mathcal{G}^1_s$, $\mathcal{G}^2_s$) in the feature binding setting. To apply supervision on $\mathcal{S}_{fb}$, $\mathcal{S}_t$, and $\mathcal{S}_p$, we upsample them to the size of $\mathcal{G}^1_s$. Then we define three pixel-wise cross-entropy losses, $\ell_{fb}$, $\ell_{t}$, and $\ell_{p}$, to measure the difference between ($\mathcal{S}_{fb}$, $\mathcal{G}^1_s$), ($\mathcal{S}_t$, $\mathcal{G}^1_s$), and ($\mathcal{S}_p$, $\mathcal{G}^2_s$), respectively. The objective function can be formalized as:
\begin{gather}
L_{stage1} = \ell_{fb} + \delta \ast \ell_{t} + (1-\delta) \ast \ell_{p} ,
\end{gather}
where $\delta$ is the weight used in to linearly combine images to generate $\mathcal{I}_{fb}$. Note that the network is penalized the most on the final and initial dominant predictions, and places less emphasis on the phantom prediction.

\subsection{Denoising Feature Binding}\label{sec:denoise}
While feature binding and source separation are interesting, the ultimate goal is to see improvement and robustness for standard images. For this reason, we mainly care about improving the overall dense prediction. To accomplish this, we further fine-tune our trained binding model on the standard training set which we call the feature denoising stage. In this stage, as we feed a standard image to the network, the phantom predictor branch, $\mathcal{F}_{ph}$, has no supervisory signal, instead it acts as a regularizer. We propose the following technique to penalize the phantom prediction.\\

\noindent \textbf{Penalize Phantom Activation:} Along with $\ell_{t}$, we propose a loss, $\ell_{\text{PPA}}$, on the phantom prediction to penalize any activation (and suppress phantom signals and interference). The goal here is to push the output of the phantom branch to zero and getting rid of the phantom. The $\ell_{\text{PPA}}$ loss sums the absolute value of the confidence attached to categories and applies a $\log$ operation to balance the numeric scale with $\ell_{t}$: 
\begin{gather}\label{eq:ppa}
\ell_{\text{PPA}} = \log\sum_{\forall_{i\in h}}\sum_{\forall_{j\in w}}\sum_{\forall_{k\in c}} \sigma (\mathcal{{S}}_{p}), \hspace{0.3cm}
L_{stage2} = \ell_{t} + \ell_{\text{PPA}},
\end{gather}
where $\sigma(\cdot)$ is the ReLU function, which constrains the input to the $\log$ to be a positive value. In \textbf{Stage 1}, $f_{\textit{enc}}$, $\mathcal{F}_t$, $\mathcal{F}_p$, and $\mathcal{F}_{fb}$ are trained in an end-to-end manner. Then, in \textbf{Stage 2}, $f_{\textit{enc}}$, $\mathcal{F}_t$, and $\mathcal{F}_p$ are fine-tuned from the Stage 1 weights.


\vspace{-0.2cm}
\section{Experiments}\label{sec:exp}
We first present results on the PASCAL VOC 2012~\cite{everingham2015pascal} semantic segmentation dataset (Sec.~\ref{sec:seg}).
Unless otherwise stated, we use the DeepLabv3~\cite{chen2017rethinking} network without any bells and whistles as our baseline model. We then show qualitative and quantitative evidence that our feature binding procedure improves the network's ability to segment highly occluded objects in complex scenes (Sec.~\ref{sec:occlusion}), as well as objects found in out-of-context scenarios (Sec.~\ref{sec:context}). Throughout the experiments, we compare our method to recent mixing strategies, CutMix~\cite{yun2019cutmix} and Mixup~\cite{zhang2017mixup}. Although Mixup and CutMix did not explicitly design their strategies for dense labeling; however, in CutMix, the authors use CutMix and MixUp for image localization and object detection tasks, so we view their strategies as a general data augmentation technique. Next, we evaluate the robustness of our method to a variety of adversarial attacks (Sec.~\ref{sec:adver}). Finally, we conduct an ablation study (Sec.~\ref{sec:ablation}) to better tease out the underlying mechanisms giving performance boosts by evaluating the various image blending strategies and network architectures.

\noindent \textbf{Implementation Details.} We implement our proposed feature binding method using Pytorch~\cite{paszke2017automatic}. 
We apply bilinear interpolation to upsample the predictions before the losses are calculated. The \textit{feature binding} network is trained using stochastic gradient descent for 30 epochs with momentum of 0.9, weight decay of 0.0005 and the “poly” learning rate policy~\cite{chen2018deeplab} which starts at $2.5e^{-4}$. We use the same strategy during the denoising stage of training, but with an initial learning rate of $2.5e^{-5}$. During training, we apply random cropping to form 321$\times$321 input images whereas testing is performed on the full resolution image.

\vspace{-0.2cm}
\subsection{Results on Semantic Segmentation}\label{sec:seg}
First, we show the improvements on segmentation accuracy by our method on the PASCAL VOC 2012 validation dataset. We present a comparison of different baselines and our proposed approach in Table~\ref{tab:voc2012_val_fcnresnet}. 
\begin{table}[t]
\begin{center}

	\begin{minipage}{0.44\linewidth}
	\begin{center}

		\setlength\tabcolsep{1.2pt}
		\def\arraystretch{1.1}
		\resizebox{0.99\textwidth}{!}{
		    \begin{tabular}{c|l|cc}
				\specialrule{1.2pt}{1pt}{1pt}
				*&\multicolumn{1}{c|}{Method}&  mIoU (\%) \\
				\specialrule{1.2pt}{1pt}{1pt}
				\multirow{4}{*}{Val}&DeepLabv3-ResNet101~\cite{chen2017rethinking} & 75.9\\
				&DeepLabv3 + Mixup~\cite{zhang2017mixup} & 75.2 \\
				&DeepLabv3 + CutMix~\cite{yun2019cutmix} & 76.0 \\
				&\textbf{DeepLabv3 + Feature Binding} & \textbf{78.0} \\

				\specialrule{1.2pt}{1pt}{1pt}
				\multirow{2}{*}{Test}&DeepLabv3~\cite{chen2017rethinking} & 79.3 \\
				&\textbf{DeepLabv3 + Feature Binding} & \textbf{82.1} \\	
				\specialrule{1.2pt}{1pt}{1pt} 

			\end{tabular}
		}
		\vspace{0.3cm}

		\caption{(a) PASCAL VOC 2012 val and test set results for the baselines and our approach.}
	    \label{tab:voc2012_val_fcnresnet}
	    	\end{center}
	\end{minipage}\hfill
	\begin{minipage}{0.52\linewidth}
	\begin{center}

		\resizebox{0.97\textwidth}{!}{
		\def\arraystretch{0.3}
        \setlength\tabcolsep{0.4pt}
		    
		\begin{tabular}{*{6}{c}}		
				
				\includegraphics[width=0.25\textwidth]{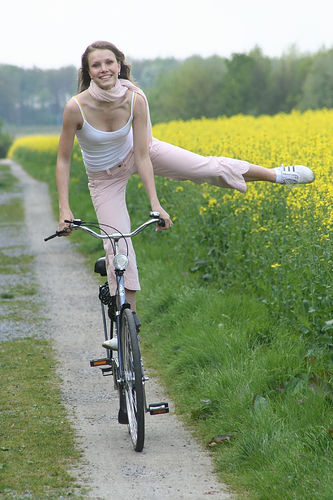}&
				\includegraphics[width=0.25\textwidth]{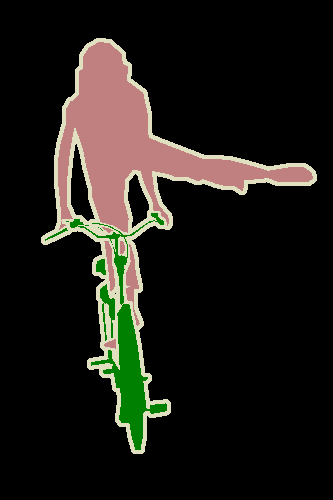}&
				\includegraphics[width=0.25\textwidth]{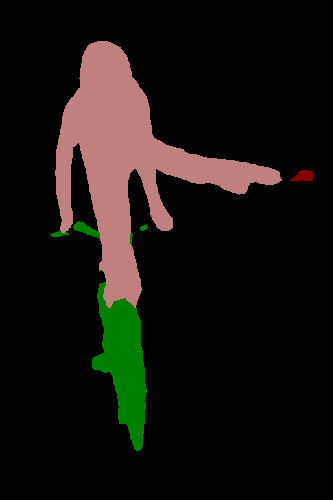}&
				\includegraphics[width=0.25\textwidth]{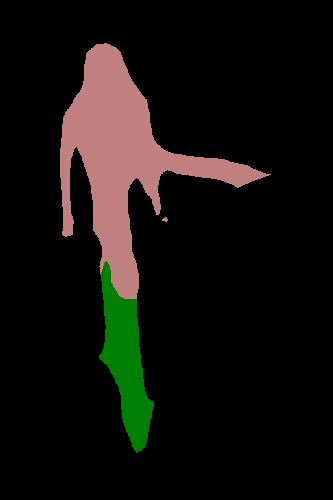}&
				\includegraphics[width=0.25\textwidth]{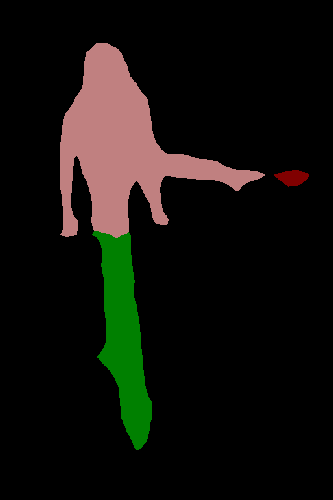}&
				\includegraphics[width=0.25\textwidth]{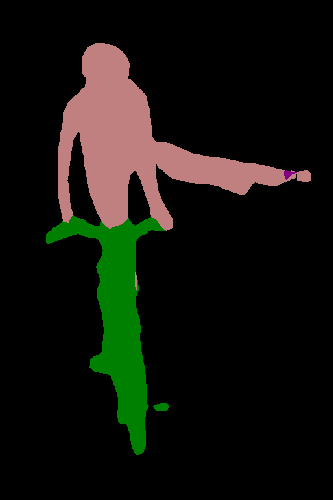}\\
				
				\includegraphics[width=0.25\textwidth]{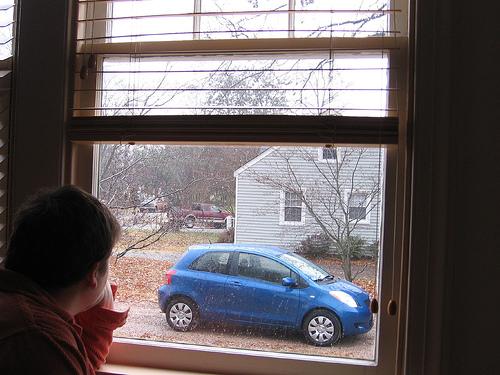}&
				\includegraphics[width=0.25\textwidth]{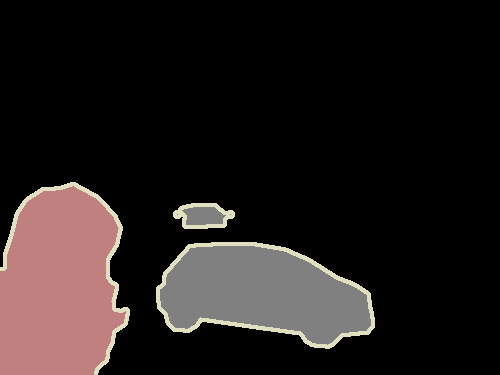}&
				\includegraphics[width=0.25\textwidth]{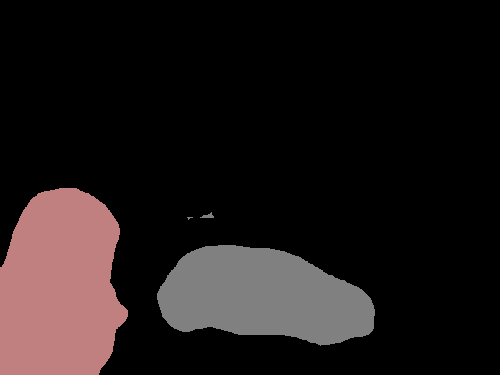}&
				\includegraphics[width=0.25\textwidth]{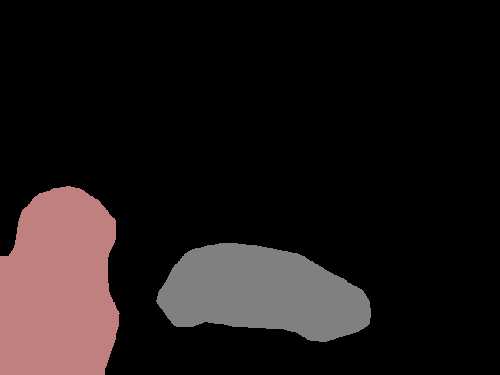}&
				\includegraphics[width=0.25\textwidth]{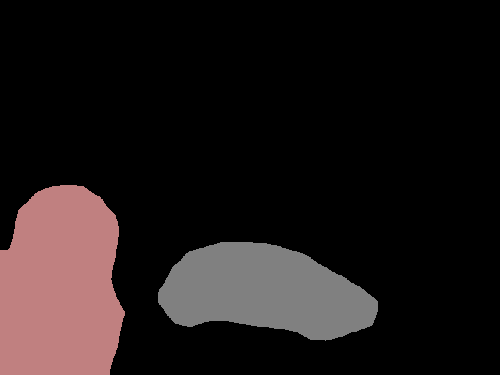}&
				\includegraphics[width=0.25\textwidth]{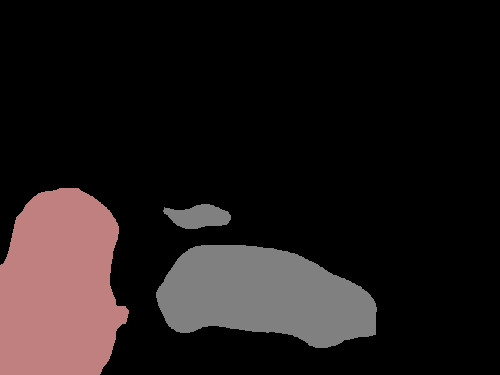}\\
				
				\includegraphics[width=0.25\textwidth]{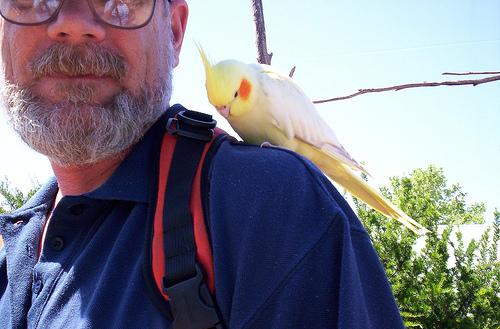}&
				\includegraphics[width=0.25\textwidth]{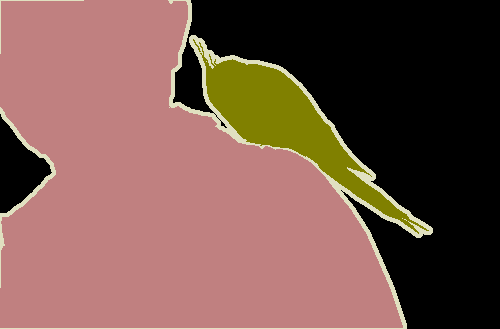}&
				\includegraphics[width=0.25\textwidth]{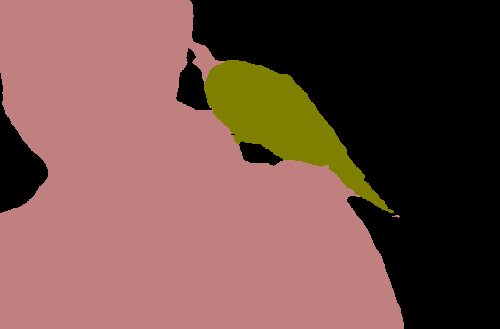}&
				\includegraphics[width=0.25\textwidth]{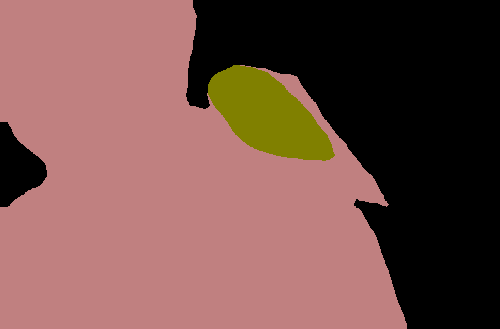}&
				\includegraphics[width=0.25\textwidth]{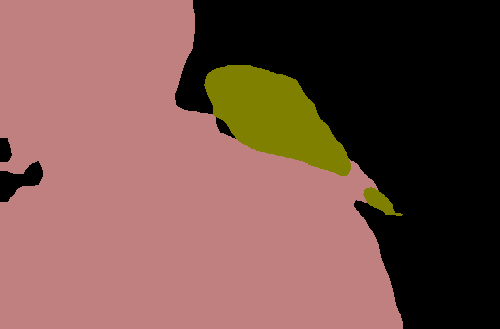}&
				\includegraphics[width=0.25\textwidth]{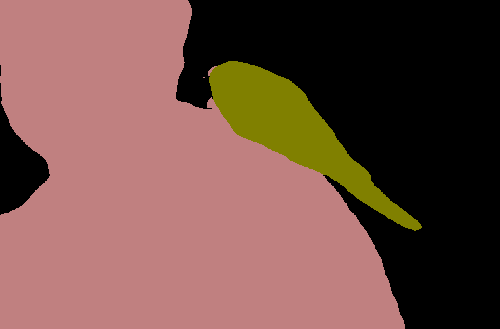}\\

				Image & GT& DeepLabv3 &CutMix& Mixup & Binding \\
			\end{tabular}}
			\vspace{0.2cm}
			\captionof{figure}{Qualitative results on the PASCAL VOC 2012 validation set.}
			\label{fig:val_pascal}
				\end{center}
	\end{minipage}
	    
\end{center}
	\vspace{-0.7cm}
\end{table}
As shown in Table~\ref{tab:voc2012_val_fcnresnet}, feature binding improves the performance significantly more than other approaches~\cite{zhang2017mixup,yun2019cutmix}. Following prior works~\cite{chen2018deeplab,zhao2017pyramid,noh15_iccv}, before evaluating our method on the test set, we first train on the augmented training set followed by fine-tuning on the original trainval set. As shown in Table~\ref{tab:voc2012_val_fcnresnet}, DeepLabv3 with feature binding achieves 82.1\% mIoU which outperforms the baseline significantly. Sample predictions of our method and the baselines are shown in Fig.~\ref{fig:val_pascal}. As shown in Fig.~\ref{fig:val_pascal}, feature binding is very effective in capturing more distinct features for labeling occluded objects and plays a critical role in separating different semantic objects more accurately. Note the ability of our method to segment scenes with a high degree of occlusion, thin overlapping regions, or complex interaction between object categories. While other methods identify the dominant categories correctly, they often fail to relate the activations of smaller occluding features to the correct categorical assignments.
\begin{table}
	\centering
	\def\arraystretch{1.0}
	\resizebox{0.98\textwidth}{!}{
		\begin{tabular}{l|cc cc ccc  c ccc}
			\specialrule{1.2pt}{1pt}{1pt}\	
			&  \multicolumn{2}{c}{\textit{Occlusion}} && \multicolumn{4}{c}{\textit{Number of Objects}}&& \multicolumn{3}{c}{\textit{Number of \textit{Unique} Objects}} \\
			\cline{2-3} \cline{5-8} \cline{10-12} 
			& \textbf{1-Occ} & \textbf{All-Occ} &&  \textbf{1-Obj} & \textbf{2-Obj
			}& \textbf{3-Obj} & \textbf{4-Obj} &&    \textbf{2-Obj}& \textbf{3-Obj} & \textbf{4-Obj}  \\
			\hline
			\hspace{0.8cm} \# of Images & 1128	& 538 && 695 & 318&	167&	98 &&		375&	121&	23 \\
			\hline
			DeepLabv3 &75.5&	74.9&&	74.6&	74.8&	76.0&	72.1&&		72.5&	\textbf{63.5}&	64.8 \\
            
            DeepLabv3 + Mixup & 75.4&	72.3&&	77.9&	74.3&	71.7&	69.6&&		72.0	&58.1&	55.4\\
           
           DeepLabv3 + CutMix & 76.4&	74.3&&	78.3&	75.4&	73.0&	72.1&&		72.3&	60.1&	56.3\\
		   
		   \textbf{DeepLabv3 + Binding} & \textbf{78.0}&	\textbf{76.2}&&	\textbf{80.7}&	\textbf{77.2}&	\textbf{76.3}&	\textbf{73.4}&&		\textbf{74.3}&	62.1&	\textbf{64.9}\\
			\specialrule{1.2pt}{1pt}{1pt}
		\end{tabular}}
		\vspace{0.3cm}
		\caption{Results on complex scenes in terms of mIoU, evaluated using various subsets from PASCAL VOC 2012 val set. \textit{Occlusion:} Number of occluded objects in the image. \textit{Number of Objects:} Number of objects in the image. \textit{Number of Unique Objects:} Unique object classes contained in the image.}
		\label{table:occlusion}
		\vspace{-0.3cm}
\end{table}
\vspace{-0.2cm}
\subsubsection{Segmenting Highly Occluded Objects in Complex Scenes}\label{sec:occlusion}
We argue that our mixing and source separation strategy is more powerful than other strategies in complex scenes with large amounts of occlusion. One reason for this is our mixing strategy (Sec.~\ref{sec:fbt}) blends images based on categorical clusters with dynamic blending ratios. This means that the network will see more images with a wide array of categories blended together, as every category is guaranteed to be blended with every other category. On the other hand, other strategies use two randomly selected images to blend. This means the statistics of the generated images will be largely driven by the statistics of the original dataset. Further, the SSM specifically is designed for separating features \textit{before} the final layer of the network, allowing for finer details and semantics to be encoded into the target and phantom streams. For the other methods, they have a single prediction, which does not allow for these details to be separated early enough in the network to encode as much information as our method.

To substantiate this claim we evaluate each method under three specific data distributions that range in amount of occlusion and complexity: (i) \textit{Occlusion}: at least one object has occlusion with any other objects (1-Occ) in an image and all objects have occlusion (All-Occ), (ii) \textit{Number of Objects}: total number of object instances regardless of classes, and (iii) \textit{Number of Unique Objects}: total number of unique semantic categories. The results are presented in Table ~\ref{table:occlusion}. Our method outperforms the other mixing based methods in all cases. Note that the improvements on all occlusion and larger number of unique categories case are particularly pronounced for our binding model as the performance drop is significantly less than the other methods, when only considering images with many unique objects.
\vspace{-0.2cm}
\subsubsection{Segmenting Out-of-Context Objects}\label{sec:context}
A model that heavily relies on context would not be able to correctly segment compared to the model that truly understands what the object is irrespective of its context. We argue that our mixing strategy performs better in out-of-context scenarios, as category-based mixing reduces bias in the dataset's co-occurrence matrix. We conduct two experiments to quantitatively evaluate each method's ability to segment out-of-context objects.

For the first experiment, we identify the top five categories that frequently co-occur with \textit{person} based on the training set, since person has the most occurrences with all other categories based on the co-occurrence matrix. We report performance in Table~\ref{table:exclusive} on two different subsets of data: (i) \textit{Co-occur with Person}: images with both the person and object in it, and (ii) \textit{Exclusive}: images with only the single object of interest. As can be seen from the table, when bottle co-occurs with person all the methods are capable of segmenting bottle and person precisely whereas the IoU for bottle is significantly reduced when bottle occurs alone. However, our proposed method successfully maintains performance on the exclusive case. 
For the second out-of-context experiment, we first create different subsets of images from the VOC 2012 val set based on the training set's co-occurrence matrix. We select thresholds $\{50, 40, 30, 20, 10\}$, and only keep images which have objects that occur less than the chosen threshold. For instance, the threshold value 50 includes all the images where the co-occurrence value of object pairs is less than 50 (e.g., cat and bottle occur 18 times together, therefore images containing both will be in all subsets except the threshold of 10). Figure~\ref{fig:frame} illustrates the result of different baselines and our method with respect to co-occurrence threshold. Our method outperforms the baselines for all the threshold values.  

\begin{table}[t]
	\centering
	\def\arraystretch{1.0}
	\resizebox{0.98\textwidth}{!}{
		\begin{tabular}{l|cc c ccc  c c ccc}
			\specialrule{1.2pt}{1pt}{1pt}\	
			
			&  \multicolumn{5}{c}{\textit{Co-occur with \textit{person}}}&& \multicolumn{5}{c}{\textit{Exclusive}} \\
			\cline{2-6} \cline{8-12} 
			& \textbf{horse} & \textbf{mbike} & \textbf{bicycle} & \textbf{bottle} & \textbf{car} & & \textbf{horse} & \textbf{mbike} & \textbf{bicycle} & \textbf{bottle} & \textbf{car} \\
			\hline
			\hspace{0.8cm} \# of Images & 32 & 34 & 30 & 20 & 45 & & 44 & 23 & 29 & 35 & 45 \\
		    \hline
			DeepLabv3 & 87.9 & 81.6 & 77.7&\textbf{89.7} & \textbf{89.7} && 90.9 & 91.5 & 60.4 & 85.4 & 96.0    \\
			DeepLabv3 + Mixup & 86.9 & 82.8 & 76.5 & 87.6 &86.2 && 92.5 & 93.0 & 60.0 & 80.6 & 95.5  \\
			DeepLabv3 + CutMix &  86.2 & 83.6 & 76.0 & 87.4 & 87.9 && \textbf{94.1} & \textbf{93.8} & 61.3 & 82.6 & 96.2\\
			\textbf{DeepLabv3 + Binding} & \textbf{90.0} & \textbf{87.7} & \textbf{79.6} & 87.9 & 89.1&& 94.0 & \textbf{93.8} & \textbf{61.9} & \textbf{88.3} & \textbf{96.6} \\
			\specialrule{1.2pt}{1pt}{1pt}
		\end{tabular}}
		\vspace{0.3cm}
		\caption{mIoU results on the PASCAL VOC 2012 val set, for the co-occurrence of the most salient person category with five other categories and the results when these five categories appear alone.}
		\label{table:exclusive}
		\vspace{-0.1cm}
\end{table}
\begin{table}[t]
	\begin{minipage}{0.44\linewidth}
       \begin{tikzpicture}
            \begin{axis}[
        xlabel={Co-occurrence threshold},
        ylabel={mIoU (\%)},
        xmin=8, xmax=65,
        ymin=45, ymax=80,
        xtick={10,20,30,40,50,60},
        xticklabels={10,20,30,40,50,Any},
        x tick label style={font=\footnotesize},
        y tick label style={font=\footnotesize},
        ytick={40,50,60,70,80},
        x dir=reverse,
        width=6cm,
        height=4.0cm,
        legend columns=1, 
        legend style={
            cells={anchor=east},
            draw=none,
            nodes={scale=0.6, transform shape},
            legend style={at={(0.8,0.565)},anchor=south}},
        ymajorgrids=true,
        x label style={at={(axis description cs:0.5,-0.1)},anchor=north,font=\footnotesize},
        y label style={at={(axis description cs:-0.077,.5)},anchor=south,font=\footnotesize},
        grid style=dashed,
    ]
    \addplot[line width=0.6pt,mark size=1.3pt,smooth,color=orange,mark=*,]
        coordinates {(60,78)(50,66.5)(40,65.6)(30,65)(20,61.6)(10,53.7)};
    \addplot[line width=0.6pt,mark size=1.2pt,smooth,color=blue,mark=square*,]
        coordinates {(60,76)(50,62.1)(40,61.4)(30,61.3)(20,57.3)(10,49.1)};
    \addplot[line width=0.6pt,mark size=1.3pt,smooth,color=red,mark=diamond*,]
        coordinates {(60,75.2)(50,64.1)(40,63.9)(30,63.8)(20,60.6)(10,51.7)};
        \legend{Binding, CutMix, Mixup}
        \end{axis}
    \end{tikzpicture}
    \vspace{0.1cm}
    \captionof{figure}{Performance on images with various levels of object co-occurence. \textit{Binding} performs better on subsets of images with unlikely co-occurrences.}
    

    \label{fig:frame}
	\end{minipage}\hfill
	\begin{minipage}{0.52\linewidth}
	\vspace{0.15cm}
	  \begin{center}
		\setlength\tabcolsep{0.7pt}
		\def\arraystretch{0.5}
		\resizebox{0.95\textwidth}{!}{
			\begin{tabular}{*{5}{c}}

				\includegraphics[width=0.32\textwidth]{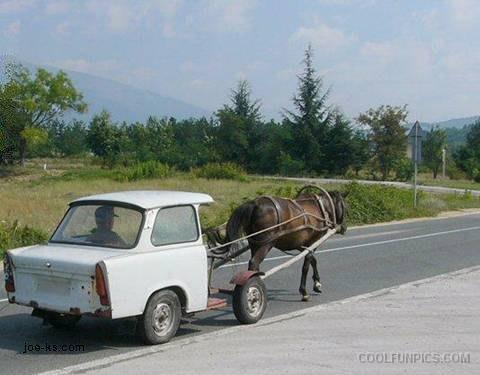}&
				\includegraphics[width=0.32\textwidth]{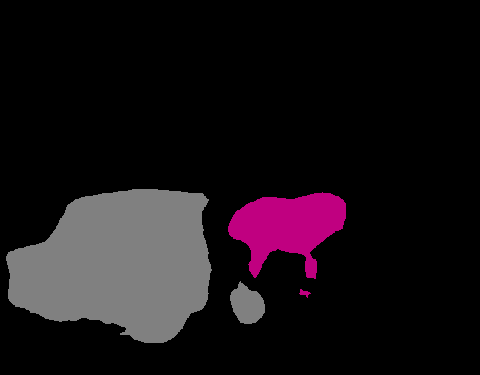}&
				\includegraphics[width=0.32\textwidth]{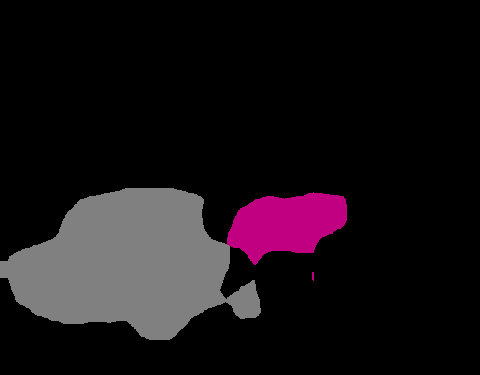}&
				\includegraphics[width=0.32\textwidth]{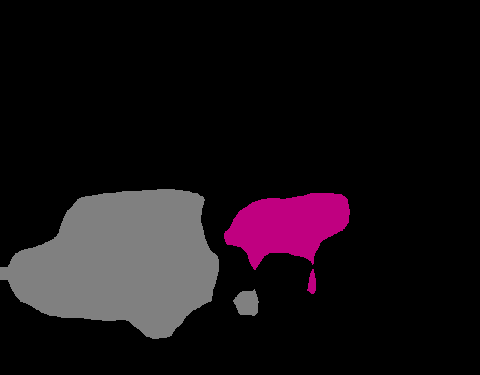}&
				\includegraphics[width=0.32\textwidth]{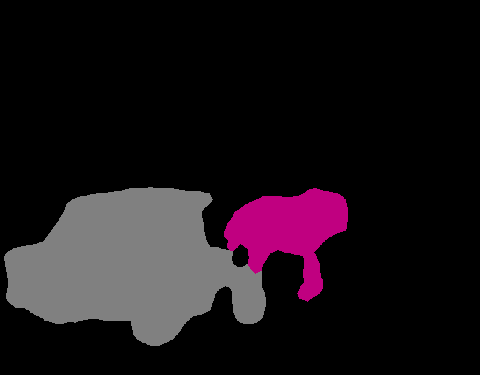}\\
				
				\includegraphics[width=0.32\textwidth, height=1.2cm]{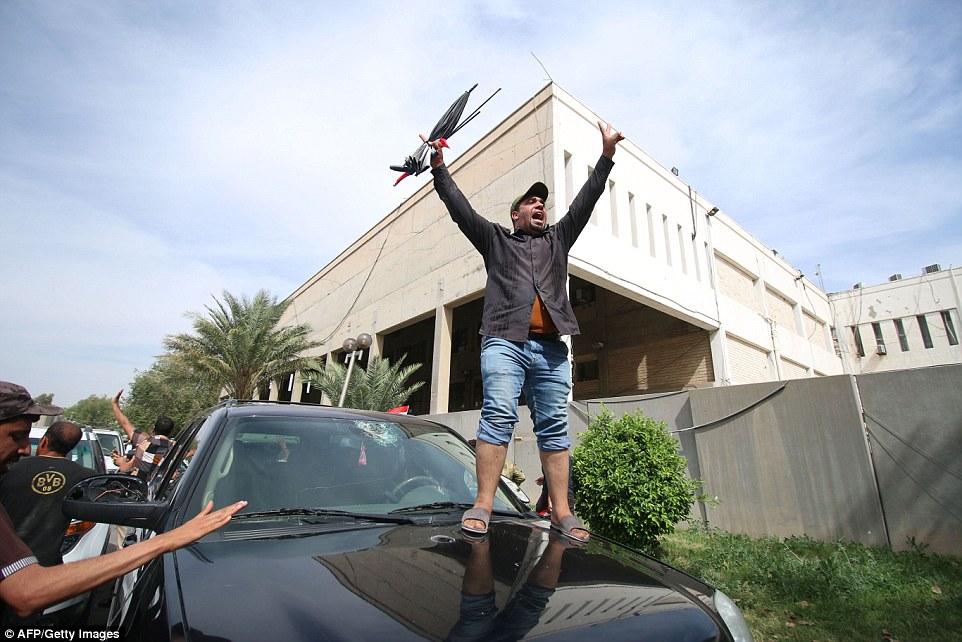}&
				\includegraphics[width=0.32\textwidth, height=1.2cm]{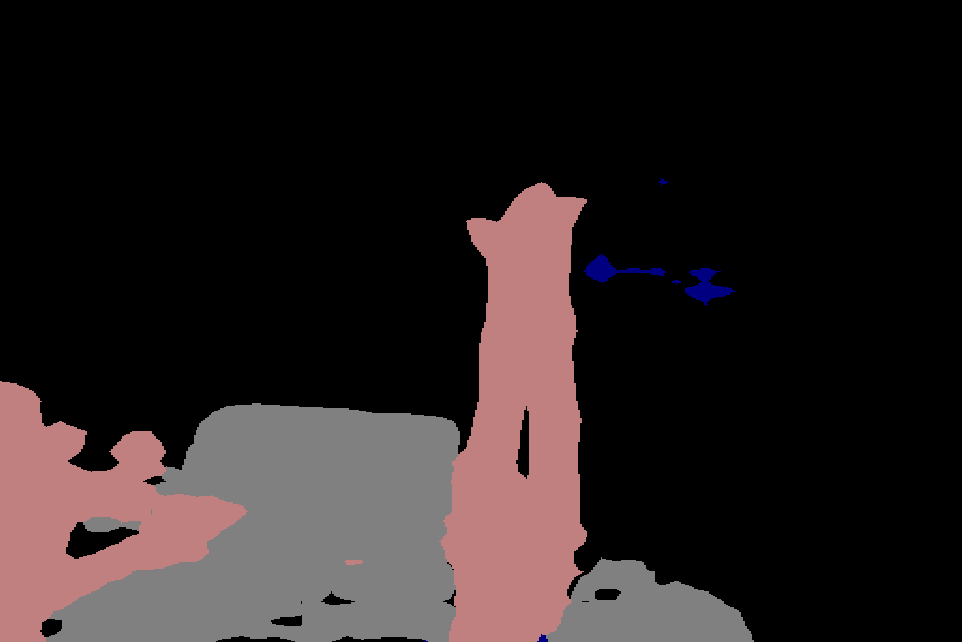}&
				\includegraphics[width=0.32\textwidth, height=1.2cm]{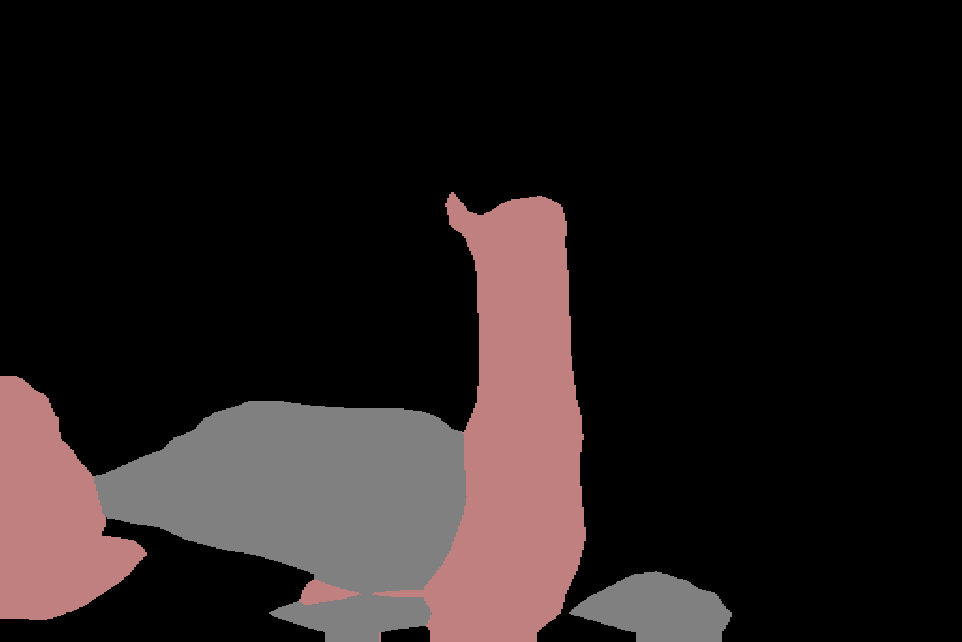}&
				\includegraphics[width=0.32\textwidth, height=1.2cm]{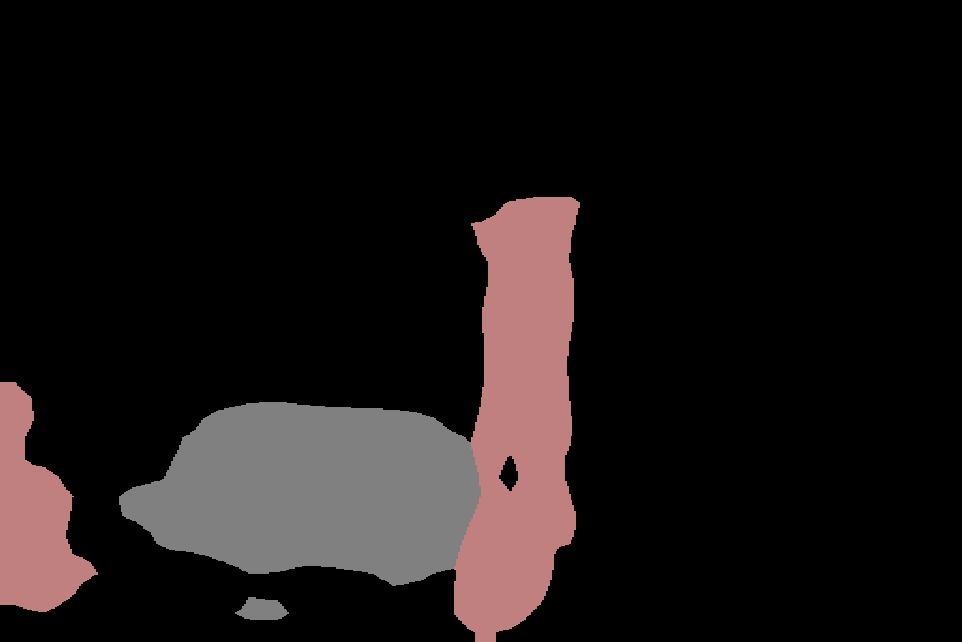}&
				\includegraphics[width=0.32\textwidth, height=1.2cm]{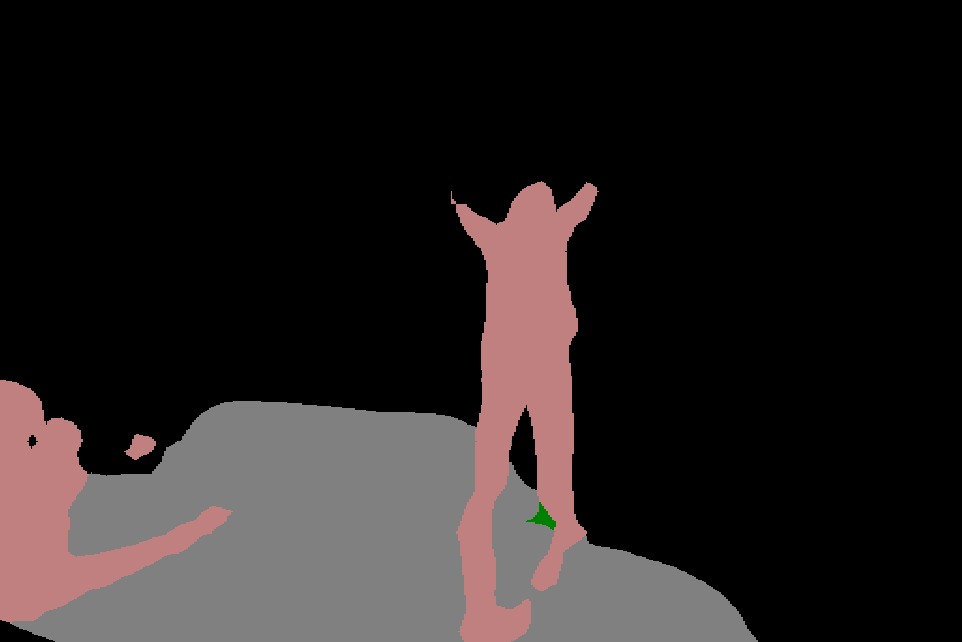}\\
				
					\includegraphics[width=0.32\textwidth]{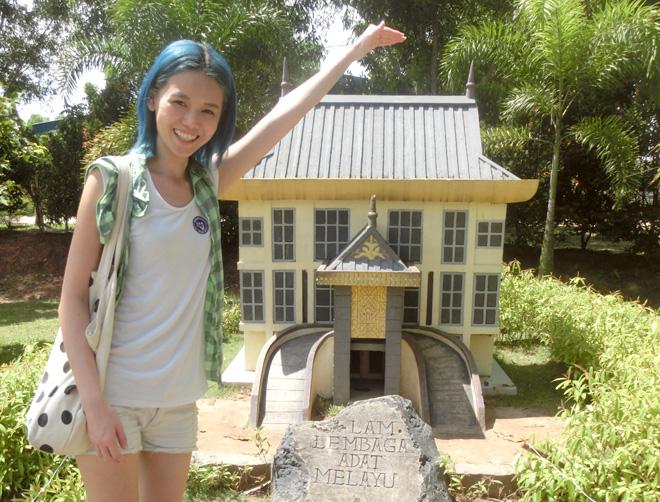}&
				\includegraphics[width=0.32\textwidth]{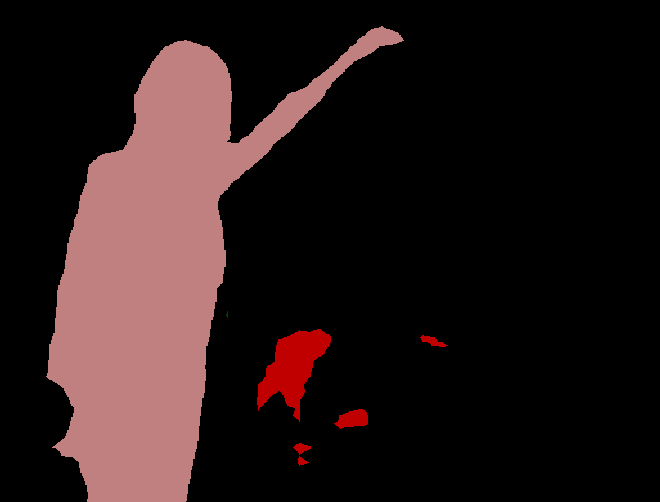}&
				\includegraphics[width=0.32\textwidth]{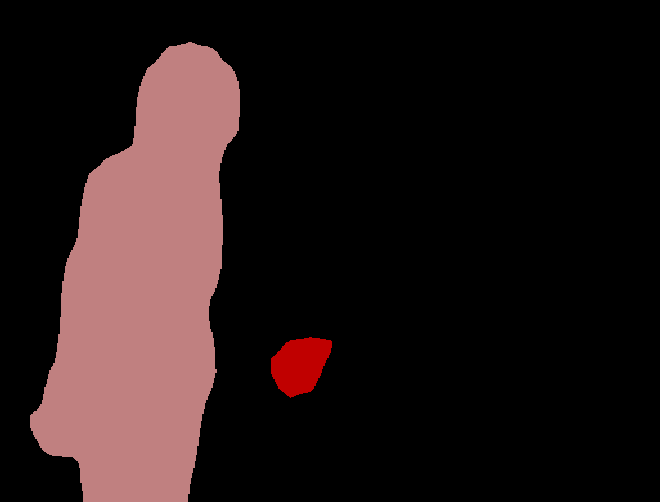}&
				\includegraphics[width=0.32\textwidth]{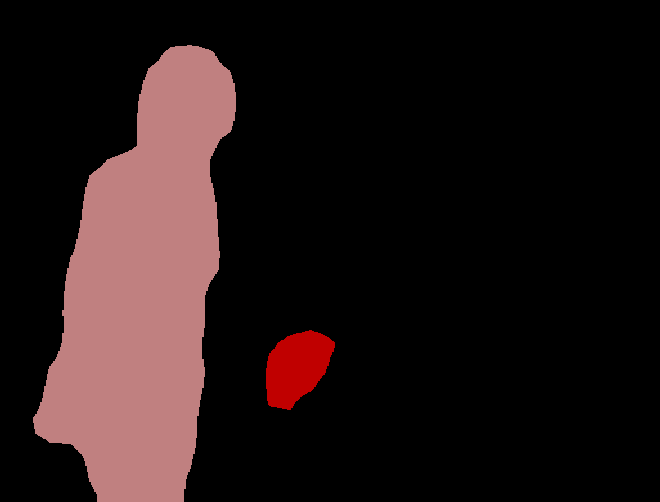}&
				\includegraphics[width=0.32\textwidth]{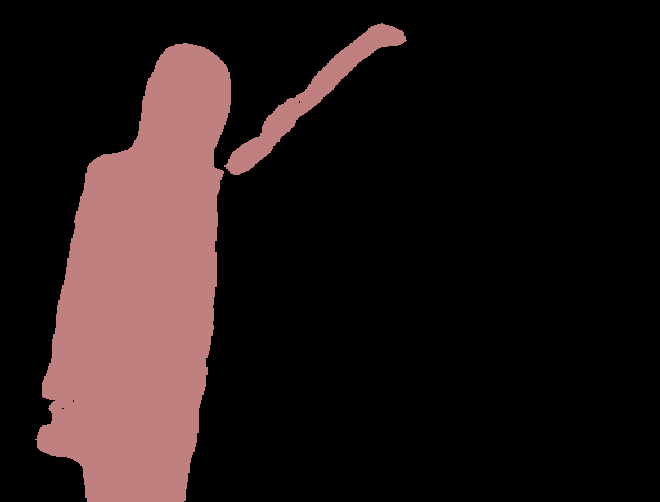}\\

				Image & DeepLabv3 &Mixup & CutMix & Ours\\
			\end{tabular}
			
			}
			\vspace{0.2cm}
			\captionof{figure}{Qualitative examples on the (top row) Out-of-Context~\cite{choi2012context} and (bottom two rows) UnRel~\cite{peyre2017weakly} datasets. Feature binding improves out-of-context performance.}
			\label{fig:valcontext}
		\end{center}
	\end{minipage}
	\vspace{-0.4cm}
\end{table}

We next perform a cross-dataset experiment by taking our model trained on the PASCAL VOC training set and evaluate on the publicly available Out-of-Context~\cite{choi2012context} and UnRel~\cite{peyre2017weakly} datasets. Fig.~\ref{fig:valcontext} visualizes how the segmentation models trained with only VOC 2012 co-occurring objects performs when objects appear without the context seen in training. Even with such challenging images with out of context objects (person \textit{on top} of car), our method produces robust segmentation masks while the baselines fails to segment the objects with detail. Since Out-of-Context and UnRel dataset do not provide segmentation ground-truth we cannot report quantitative results on these datasets. 
\subsection{Adversarial Robustness}\label{sec:adver}
We further claim our technique works as an implicit defense mechanism against adversarial images similar to~\cite{yun2019cutmix,zhang2017mixup,inoue2018data,cubuk2019autoaugment,harris2020fmix,chou2020remix}. This is because the network optimization, in the form of source separation to solve the binding problem, enhances the capability of interacting with noisy features while imposing a high degree of resilience to interference from the superimposed image.\\
\begin{table}[t]
\vspace{0.1cm}
	\begin{minipage}{0.55\linewidth}
     \centering
	\def\arraystretch{1.0}
	\resizebox{0.98\textwidth}{!}{
		\begin{tabular}{l|c|cc c ccc}
			\specialrule{1.2pt}{1pt}{1pt}\	
			
			\multirow{3}{*}{\hspace{0.1cm} Networks}& \multirow{3}{*}{Clean} & \multicolumn{5}{c}{\textbf{Adversarial Images}}\\
			\cline{3-8}
			& & \multicolumn{2}{c}{UAP~\cite{moosavi2017universal}}&& \multicolumn{3}{c}{GD-UAP~\cite{mopuri2018generalizable}}  \\
			
			\cline{3-4} \cline{6-8}
			
			&&ResNet&GNet& &R-No&R-All &R-Part \\
			
			\hline
			\hline
			
			DeepLabv3 & 75.9&59.1 & 63.6 && 58.7&56.8 &  56.5   \\
			+ Mixup & 75.2 & 62.9& 63.2 && 54.8&53.0 & 53.6    \\
			 + CutMix & 76.2 & 60.9& 64.3 && 47.4&46.4 & 46.9   \\
			 + \textbf{Binding} & \textbf{78.0} &\textbf{68.7} & \textbf{70.2}&&  \textbf{63.9}& \textbf{63.1} & \textbf{62.8}   \\	
			
			\specialrule{1.2pt}{1pt}{1pt}
			
		\end{tabular}}
		\vspace{0.3cm}
		\caption{The mean IoU for baselines and our approach when attacked with UAP~\cite{moosavi2017universal} and GD-UAP~\cite{mopuri2018generalizable}.}
		\label{table:noisy-analysis}
	\end{minipage}\hfill
	\begin{minipage}{0.43\linewidth}
	  \begin{center}
		\setlength\tabcolsep{0.3pt}
		\def\arraystretch{0.3}
		\resizebox{0.95\textwidth}{!}{
			\begin{tabular}{*{4}{c}}		

				\includegraphics[width=0.30\textwidth]{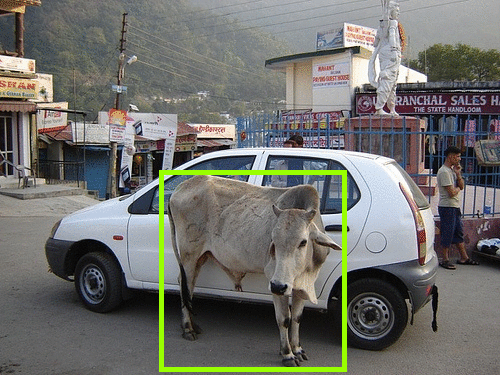}&
				\includegraphics[width=0.30\textwidth]{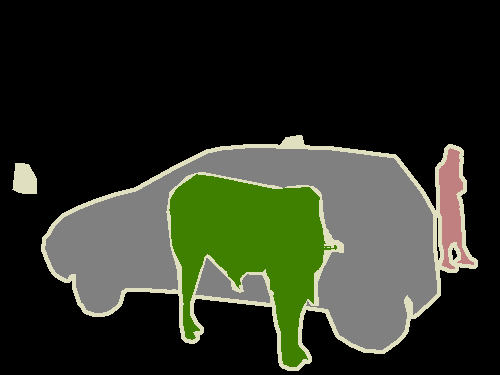}&
				\includegraphics[width=0.30\textwidth]{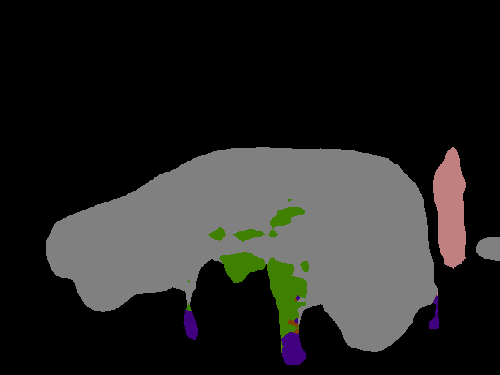}&
				\includegraphics[width=0.30\textwidth]{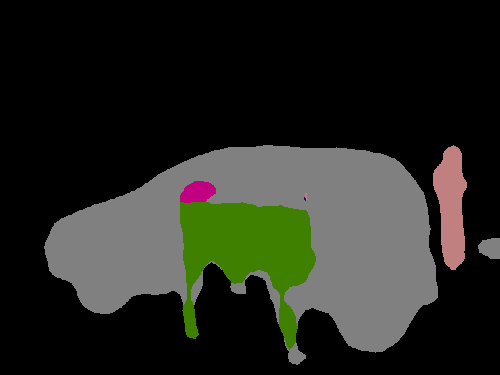}\\		
				
				\includegraphics[width=0.30\textwidth]{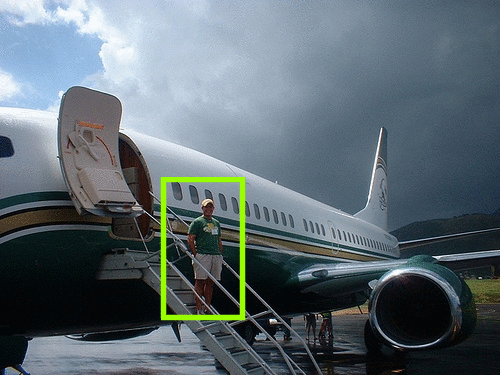}&
				\includegraphics[width=0.30\textwidth]{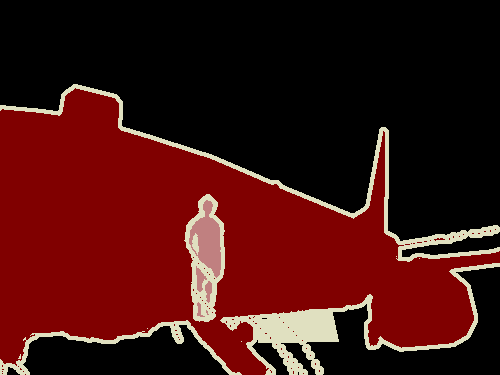}&
				\includegraphics[width=0.30\textwidth]{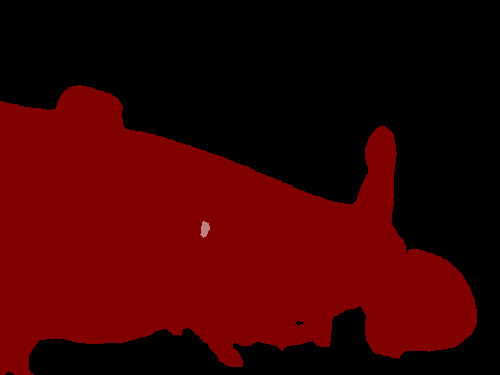}&
				\includegraphics[width=0.30\textwidth]{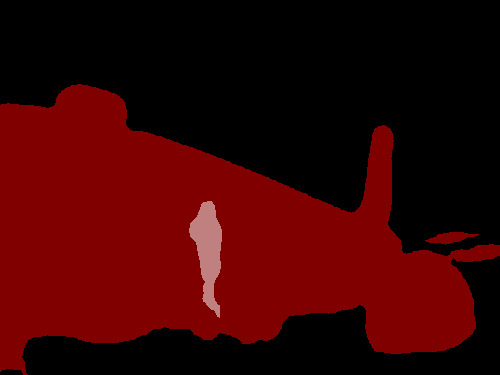}\\

				Image & GT& DeepLabv3 & Binding
			\end{tabular}
			}
			\vspace{0.2cm}
			\captionof{figure}{Two challenging images where semantic objects are highly occluded.}
			\label{fig:discuss}
		\end{center}

	\end{minipage}
	\vspace{-0.2cm}
\end{table}

\noindent \textbf{Adversarial Attacks.} We generate adversarial examples using various techniques, including the Universal Adversarial Perturbation (UAP)~\cite{moosavi2017universal} and Generalizable Data-free Universal Adversarial Perturbation (GD-UAP)~\cite{mopuri2018generalizable} under different settings. We use publicly available perturbations of these methods to generate adversarial examples for the VOC 2012 val set. For UAP, which is a black-box attack, we generate adversarial images with both ResNet152 and GoogleNet based universal perturbations. GD-UAP is a grey-box attack, as it generates a perturbation based on the source data (VOC 2012 train set) and the backbone network (ResNet101). For GD-UAP, we compare different levels of adversarial attack strength by generating the perturbation based on various amounts of source data information.\\

\noindent \textbf{Robustness of Segmentation Networks.} We evaluate the robustness of different methods to adversarial examples and show how feature binding-driven training learns to significantly mitigate performance loss due to perturbation. Table~\ref{table:noisy-analysis} shows the robustness of different baselines and our approach on the PASCAL VOC 2012 validation dataset. In general, DeepLab-based methods~\cite{chen2018deeplab} achieve higher mIoU for the segmentation task on clean examples and is also shown to be more robust to adversarial samples compared to the shallower networks~\cite{arnab2018robustness}. In the case of black-box attacks, the adversarial examples originally generated by UAP on ResNet152, are less malignant (68.7\% mIoU) when the feature binding concept is applied, while being effective in significantly reducing the performance of other methods.  

When we apply a semi-white-box attack under the setting (R-All), where VOC 2012 training data and the ResNet101 network are used to generate the perturbation, DeepLabv3 and Mixup show robustness against adversarial examples which is improved by applying feature binding. Surprisingly, the performance of CutMix is significantly reduced when tested against adversarial samples generated by GD-UAP. Similarly, we find that DeepLabv3, Mixup, and CutMix are also vulnerable to adversarial cases under the \textit{R-No} and \textit{R-Part} settings, where no data and partial data is used respectively to generate the perturbations. Notably, DeepLabv3+Binding exhibits significant robustness to extreme cases which further reveals the importance of feature binding to successfully relate internal activations corresponding to common sources in the adversarial images. \\
\vspace{-0.7cm}
\subsection{Ablation Studies}\label{sec:ablation}
\vspace{-0.1cm}
In this section, we examine the possible variants of our feature binding pipeline by considering three different settings. Note that for all the experiments, except the denoising in our ablation study, we choose ResNet101 based distributed gating network~\cite{karim2020distributed} as the backbone. \\

\noindent \textbf{Feature Binding Driven Blending Techniques.} 
We report the labeling results of several blending techniques in Table~\ref{tab:fb_quan}(a). If we select two images randomly and allow the two images to be any class ($\mathcal{R}_{f_b}$), the performance is lower than the proposed clustering based technique. Additionally, the performance was not improved when we mix two images belonging to the same category ($\mathcal{C}a_{f_b}$). However, our proposed clustering based blending, $\mathcal{C}_{f_b}$, achieves higher mIoU compared to possible alternatives highlighting the importance of the choice of pattern collisions in applying feature binding.\\

\noindent \textbf{Feature Denoising and Feature Binding Head.} We examine the effectiveness of feature denoising (DN) stage and report results in Table~\ref{tab:fb_quan}(b). We also conduct experiments (Table~\ref{tab:fb_quan}(c)) varying the source separator module, including the feature binding head (FBH). The overall performance can be improved with the addition of a feature denoising stage and feature binding head, see Table~\ref{tab:fb_quan} (b) and (c), respectively. We believe the feature binding head allows the network to make a more informed final prediction based on the source \textit{and} the phantom activations, and therefore learns to identify harmful features at inference time, leading to a more accurate prediction. 
\begin{table}
	\begin{center}
		\def\arraystretch{1.0}
		\setlength\tabcolsep{1.3pt}
		\resizebox{0.44\textwidth}{!}{
			\begin{tabular}{l|cccccc}
				\specialrule{1.2pt}{1pt}{1pt}
				\multirow{2}{*}{Method}& \multicolumn{6}{c}{\textbf{Image Blending Techniques}}\\
				\cline{2-7}
				&No&$\mathcal{C}_{f_b}$ &$\mathcal{R}_{f_b}$ &$\mathcal{C}a_{f_b}$ & $W\mathcal{R}_{f_b}$ &  $\mathcal{M}_{f_b}$    \\
				\specialrule{1.2pt}{1pt}{1pt}
				DIGNet~\cite{karim2020distributed} &75.1&\textbf{76.1}&74.5 &74.5&69.7&75.4\\
				\specialrule{1.2pt}{1pt}{1pt}
				\multicolumn{7}{c}{(a)}

		\end{tabular}}
		\hspace{0.1cm}
		\resizebox{0.23\textwidth}{!}{
			\begin{tabular}{l|c}
				\specialrule{1.2pt}{1pt}{1pt}
				
				\multicolumn{1}{c|}{Methods} & mIoU \\
				\hline
				DeepLabv3 & 75.9 \\
				+ ours (w/o DN) & 76.2 \\
				+ ours (w/ DN) & \textbf{78.0} \\

				\specialrule{1.2pt}{1pt}{1pt}
				\multicolumn{2}{c}{(b)}
			\end{tabular}}
		\hspace{0.1cm}
		\resizebox{0.24\textwidth}{!}{
		    \begin{tabular}{l|c}
				\specialrule{1.2pt}{1pt}{1pt}
				
				\multicolumn{1}{c|}{Methods} & mIoU \\
				\hline
				DIGNet~\cite{karim2020distributed} & 75.1 \\
				+ Ours (w/o FBH) & 75.3 \\
				+ Ours (w/ FBH) & \textbf{76.1} \\

				\specialrule{1.2pt}{1pt}{1pt}
				\multicolumn{2}{c}{(c)}
			\end{tabular}
		}
		\vspace{0.2cm}
		\caption{(a) Performance comparison of different blending techniques on the VOC 2012 val set. $\mathcal{C}_{f_b}$: Clustering based blending discussed in Sec.~\ref{sec:fbt},  $\mathcal{R}_{f_b}$: Each sample randomly paired with 10 samples, $\mathcal{C}a_{f_b}$: Within category random pair, $W\mathcal{R}_{f_b}$: random pairing with fixed $\delta=0.6$, $\mathcal{M}_{f_b}$: Random pairs from half of the train set and using standard images from the other half. (b) Significance of feature denoising stage (DN). (c) Performance comparison with and without the feature binding head (FBH).}
		\label{tab:fb_quan}
	\end{center}
    \vspace{-0.5cm}
\end{table}

\vspace{-0.3cm}
\section{Discussion and Conclusion}\label{sec:conclude}
Training with the feature binding pipeline enables learning resilient features, separating sources of activation, and resolving ambiguity with richer contextual information. Although DeepLabv3 is a powerful segmentation network, there are cases (see Fig.~\ref{fig:discuss}) where background objects are correctly classified (car and plane) but other semantic categories are not separated correctly due to high degrees of occlusion (person on the stairs, see Fig.~\ref{fig:discuss} right). In contrast, the feature binding based learning approach is highly capable of resolving such cases by learning to separate source objects and tying them to specific regions.

In summary, we have presented an approach to training CNNs based on the notion of feature binding. This process includes, as one major component, careful creation of categorical collisions in data during training. This results in improved segmentation performance, and also promotes significant robustness to adversarial perturbations. Denoising in the form of fine-tuning shows further improvement along both these dimensions. 



\vspace{0.2cm}
\noindent \textbf{Acknowledgements:} The authors gratefully acknowledge financial support from the Canadian NSERC Discovery Grants, Ontario Graduate Scholarship, and Vector Institute Postgraduate Affiliation award.  K.G.D. contributed to this work in his personal capacity as an Associate Professor at Ryerson University. We also thank the NVIDIA Corporation for providing GPUs through their academic program.

\bibliography{paper,paper2,paper_3}

\begin{thebibliography}{40}
\providecommand{\natexlab}[1]{#1}
\providecommand{\url}[1]{\texttt{#1}}
\expandafter\ifx\csname urlstyle\endcsname\relax
  \providecommand{\doi}[1]{doi: #1}\else
  \providecommand{\doi}{doi: \begingroup \urlstyle{rm}\Url}\fi

\bibitem[Arnab et~al.(2018)Arnab, Miksik, and Torr]{arnab2018robustness}
Anurag Arnab, Ondrej Miksik, and Philip~HS Torr.
\newblock On the robustness of semantic segmentation models to adversarial
  attacks.
\newblock In \emph{CVPR}, 2018.

\bibitem[Badrinarayanan et~al.(2017)Badrinarayanan, Kendall, and
  Cipolla]{badrinarayanan15_arxiv}
Vijay Badrinarayanan, Alex Kendall, and Roberto Cipolla.
\newblock Segnet: A deep convolutional encoder-decoder architecture for scene
  segmentation.
\newblock \emph{TPAMI}, 2017.

\bibitem[Chen et~al.(2015)Chen, Papandreou, Kokkinos, Murphy, and
  Yuille]{chen15_iclr}
Liang-Chieh Chen, George Papandreou, Iasonas Kokkinos, Kevin Murphy, and
  Alan~L. Yuille.
\newblock Semantic image segmentation with deep convolutional nets and fully
  connected {CRF}s.
\newblock In \emph{ICLR}, 2015.

\bibitem[Chen et~al.(2017)Chen, Papandreou, Schroff, and
  Adam]{chen2017rethinking}
Liang-Chieh Chen, George Papandreou, Florian Schroff, and Hartwig Adam.
\newblock Rethinking atrous convolution for semantic image segmentation.
\newblock \emph{arXiv:1706.05587}, 2017.

\bibitem[Chen et~al.(2018)Chen, Papandreou, Kokkinos, Murphy, and
  Yuille]{chen2018deeplab}
Liang-Chieh Chen, George Papandreou, Iasonas Kokkinos, Kevin Murphy, and Alan~L
  Yuille.
\newblock Deeplab: Semantic image segmentation with deep convolutional nets,
  atrous convolution, and fully connected crfs.
\newblock \emph{TPAMI}, 2018.

\bibitem[Choi et~al.(2012)Choi, Torralba, and Willsky]{choi2012context}
Myung~Jin Choi, Antonio Torralba, and Alan~S Willsky.
\newblock Context models and out-of-context objects.
\newblock \emph{Pattern Recognition Letters}, 2012.

\bibitem[Chou et~al.(2020)Chou, Chang, Pan, Wei, and Juan]{chou2020remix}
Hsin-Ping Chou, Shih-Chieh Chang, Jia-Yu Pan, Wei Wei, and Da-Cheng Juan.
\newblock Remix: Rebalanced mixup.
\newblock \emph{arXiv preprint arXiv:2007.03943}, 2020.

\bibitem[Cubuk et~al.(2019)Cubuk, Zoph, Mane, Vasudevan, and
  Le]{cubuk2019autoaugment}
Ekin~D Cubuk, Barret Zoph, Dandelion Mane, Vijay Vasudevan, and Quoc~V Le.
\newblock Autoaugment: Learning augmentation strategies from data.
\newblock In \emph{CVPR}, 2019.

\bibitem[DeVries and Taylor(2017)]{devries2017improved}
Terrance DeVries and Graham~W Taylor.
\newblock Improved regularization of convolutional neural networks with cutout.
\newblock \emph{arXiv preprint arXiv:1708.04552}, 2017.

\bibitem[Everingham et~al.(2015)Everingham, Eslami, Van~Gool, Williams, Winn,
  and Zisserman]{everingham2015pascal}
Mark Everingham, SM~Ali Eslami, Luc Van~Gool, Christopher~KI Williams, John
  Winn, and Andrew Zisserman.
\newblock The pascal visual object classes challenge: A retrospective.
\newblock \emph{IJCV}, 2015.

\bibitem[French et~al.(2019)French, Aila, Laine, Mackiewicz, and
  Finlayson]{french2019semi}
Geoff French, Timo Aila, Samuli Laine, Michal Mackiewicz, and Graham Finlayson.
\newblock Semi-supervised semantic segmentation needs strong, high-dimensional
  perturbations.
\newblock \emph{arXiv preprint arXiv:1906.01916}, 2019.

\bibitem[Georgiev et~al.(2005)Georgiev, Theis, and
  Cichocki]{georgiev2005sparse}
Pando Georgiev, Fabian Theis, and Andrzej Cichocki.
\newblock Sparse component analysis and blind source separation of
  underdetermined mixtures.
\newblock \emph{TNN}, 16\penalty0 (4):\penalty0 992--996, 2005.

\bibitem[Ghiasi and Fowlkes(2016)]{ghiasi2016laplacian}
Golnaz Ghiasi and Charless~C Fowlkes.
\newblock Laplacian pyramid reconstruction and refinement for semantic
  segmentation.
\newblock In \emph{ECCV}, 2016.

\bibitem[Harris et~al.(2020)Harris, Marcu, Painter, Niranjan, and
  Hare]{harris2020fmix}
Ethan Harris, Antonia Marcu, Matthew Painter, Mahesan Niranjan, and Adam
  Pr{\"u}gel-Bennett~Jonathon Hare.
\newblock Fmix: Enhancing mixed sample data augmentation.
\newblock \emph{arXiv preprint arXiv:2002.12047}, 2020.

\bibitem[He et~al.(2017)He, Gkioxari, Doll{\'a}r, and Girshick]{he2017mask}
Kaiming He, Georgia Gkioxari, Piotr Doll{\'a}r, and Ross Girshick.
\newblock Mask r-cnn.
\newblock In \emph{ICCV}, 2017.

\bibitem[Huang et~al.(2015)Huang, Kim, Hasegawa-Johnson, and
  Smaragdis]{huang2015joint}
Po-Sen Huang, Minje Kim, Mark Hasegawa-Johnson, and Paris Smaragdis.
\newblock Joint optimization of masks and deep recurrent neural networks for
  monaural source separation.
\newblock \emph{ASLP}, 2015.

\bibitem[Inoue(2018)]{inoue2018data}
Hiroshi Inoue.
\newblock Data augmentation by pairing samples for images classification.
\newblock \emph{arXiv:1801.02929}, 2018.

\bibitem[Islam et~al.(2017{\natexlab{a}})Islam, Naha, Rochan, Bruce, and
  Wang]{islam2017label}
Md~Amirul Islam, Shujon Naha, Mrigank Rochan, Neil Bruce, and Yang Wang.
\newblock Label refinement network for coarse-to-fine semantic segmentation.
\newblock \emph{arXiv:1703.00551}, 2017{\natexlab{a}}.

\bibitem[Islam et~al.(2017{\natexlab{b}})Islam, Rochan, Bruce, and
  Wang]{Islam_2017_CVPR}
Md~Amirul Islam, Mrigank Rochan, Neil D.~B. Bruce, and Yang Wang.
\newblock Gated feedback refinement network for dense image labeling.
\newblock In \emph{CVPR}, 2017{\natexlab{b}}.

\bibitem[Islam et~al.(2018{\natexlab{a}})Islam, Kalash, and Bruce]{cvpr18_rank}
Md~Amirul Islam, Mahmoud Kalash, and Neil~D.B. Bruce.
\newblock Revisiting salient object detection: Simultaneous detection, ranking,
  and subitizing of multiple salient objects.
\newblock In \emph{CVPR}, 2018{\natexlab{a}}.

\bibitem[Islam et~al.(2018{\natexlab{b}})Islam, Kalash, and
  Bruce]{islam2018semantics}
Md~Amirul Islam, Mahmoud Kalash, and Neil~DB Bruce.
\newblock Semantics meet saliency: Exploring domain affinity and models for
  dual-task prediction.
\newblock In \emph{BMVC}, 2018{\natexlab{b}}.

\bibitem[Islam et~al.(2018{\natexlab{c}})Islam, Rochan, Naha, Bruce, and
  Wang]{islam2018gated}
Md~Amirul Islam, Mrigank Rochan, Shujon Naha, Neil~DB Bruce, and Yang Wang.
\newblock Gated feedback refinement network for coarse-to-fine dense semantic
  image labeling.
\newblock \emph{arXiv preprint arXiv:1806.11266}, 2018{\natexlab{c}}.

\bibitem[Karim et~al.(2019)Karim, Islam, and Bruce]{karim2019recurrent}
Rezaul Karim, Md~Amirul Islam, and Neil~DB Bruce.
\newblock Recurrent iterative gating networks for semantic segmentation.
\newblock In \emph{WACV}, 2019.

\bibitem[Karim et~al.(2020)Karim, Islam, and Bruce]{karim2020distributed}
Rezaul Karim, Md~Amirul Islam, and Neil~DB Bruce.
\newblock Distributed iterative gating networks for semantic segmentation.
\newblock In \emph{WACV}, 2020.

\bibitem[Li et~al.(2016)Li, Hariharan, and Malik]{li2016iterative}
Ke~Li, Bharath Hariharan, and Jitendra Malik.
\newblock Iterative instance segmentation.
\newblock In \emph{CVPR}, 2016.

\bibitem[Lin et~al.(2014)Lin, Maire, Belongie, Hays, Perona, Ramanan,
  Doll{\'a}r, and Zitnick]{lin2014microsoft}
Tsung-Yi Lin, Michael Maire, Serge Belongie, James Hays, Pietro Perona, Deva
  Ramanan, Piotr Doll{\'a}r, and C~Lawrence Zitnick.
\newblock Microsoft coco: Common objects in context.
\newblock In \emph{ECCV}, 2014.

\bibitem[Long et~al.(2015)Long, Shelhamer, and Darrell]{long15_cvpr}
Jonathan Long, Evan Shelhamer, and Trevor Darrell.
\newblock Fully convolutional networks for semantic segmentation.
\newblock In \emph{CVPR}, 2015.

\bibitem[Moosavi-Dezfooli et~al.(2017)Moosavi-Dezfooli, Fawzi, Fawzi, and
  Frossard]{moosavi2017universal}
Seyed-Mohsen Moosavi-Dezfooli, Alhussein Fawzi, Omar Fawzi, and Pascal
  Frossard.
\newblock Universal adversarial perturbations.
\newblock In \emph{CVPR}, 2017.

\bibitem[Mopuri et~al.(2018)Mopuri, Ganeshan, and
  Babu]{mopuri2018generalizable}
Konda~Reddy Mopuri, Aditya Ganeshan, and R~Venkatesh Babu.
\newblock Generalizable data-free objective for crafting universal adversarial
  perturbations.
\newblock \emph{TPAMI}, 2018.

\bibitem[Noh et~al.(2015)Noh, Hong, and Han]{noh15_iccv}
Hyeonwoo Noh, Seunghoon Hong, and Bohyung Han.
\newblock Learning deconvolution network for semantic segmentation.
\newblock In \emph{ICCV}, 2015.

\bibitem[Paszke et~al.(2017)Paszke, Gross, Chintala, Chanan, Yang, DeVito, Lin,
  Desmaison, Antiga, and Lerer]{paszke2017automatic}
Adam Paszke, Sam Gross, Soumith Chintala, Gregory Chanan, Edward Yang, Zachary
  DeVito, Zeming Lin, Alban Desmaison, Luca Antiga, and Adam Lerer.
\newblock Automatic differentiation in pytorch.
\newblock 2017.

\bibitem[Peyre et~al.(2017)Peyre, Sivic, Laptev, and Schmid]{peyre2017weakly}
Julia Peyre, Josef Sivic, Ivan Laptev, and Cordelia Schmid.
\newblock Weakly-supervised learning of visual relations.
\newblock In \emph{ICCV}, 2017.

\bibitem[Russakovsky et~al.(2015)Russakovsky, Deng, Su, Krause, Satheesh, Ma,
  Huang, Karpathy, Khosla, Bernstein, et~al.]{russakovsky2015imagenet}
Olga Russakovsky, Jia Deng, Hao Su, Jonathan Krause, Sanjeev Satheesh, Sean Ma,
  Zhiheng Huang, Andrej Karpathy, Aditya Khosla, Michael Bernstein, et~al.
\newblock Imagenet large scale visual recognition challenge.
\newblock \emph{IJCV}, 2015.

\bibitem[Shipp et~al.(2009)Shipp, Adams, Moutoussis, and
  Zeki]{shipp2009feature}
Stewart Shipp, Daniel~L Adams, Konstantinos Moutoussis, and Semir Zeki.
\newblock Feature binding in the feedback layers of area v2.
\newblock \emph{Cerebral cortex}, 19\penalty0 (10):\penalty0 2230--2239, 2009.

\bibitem[Singh et~al.(2020)Singh, Mahajan, Grauman, Lee, Feiszli, and
  Ghadiyaram]{singh2020don}
Krishna~Kumar Singh, Dhruv Mahajan, Kristen Grauman, Yong~Jae Lee, Matt
  Feiszli, and Deepti Ghadiyaram.
\newblock Don't judge an object by its context: Learning to overcome contextual
  bias.
\newblock \emph{arXiv preprint arXiv:2001.03152}, 2020.

\bibitem[Tokozume et~al.(2018)Tokozume, Ushiku, and
  Harada]{tokozume2018between}
Yuji Tokozume, Yoshitaka Ushiku, and Tatsuya Harada.
\newblock Between-class learning for image classification.
\newblock In \emph{CVPR}, 2018.

\bibitem[Treisman(1998)]{treisman1998feature}
Anne Treisman.
\newblock Feature binding, attention and object perception.
\newblock \emph{Phil. Trans. R. Soc. B.}, 1998.

\bibitem[Yun et~al.(2019)Yun, Han, Oh, Chun, Choe, and Yoo]{yun2019cutmix}
Sangdoo Yun, Dongyoon Han, Seong~Joon Oh, Sanghyuk Chun, Junsuk Choe, and
  Youngjoon Yoo.
\newblock Cutmix: Regularization strategy to train strong classifiers with
  localizable features.
\newblock In \emph{ICCV}, 2019.

\bibitem[Zhang et~al.(2018)Zhang, Cisse, Dauphin, and
  Lopez-Paz]{zhang2017mixup}
Hongyi Zhang, Moustapha Cisse, Yann~N Dauphin, and David Lopez-Paz.
\newblock mixup: Beyond empirical risk minimization.
\newblock In \emph{ICLR}, 2018.

\bibitem[Zhao et~al.(2017)Zhao, Shi, Qi, Wang, and Jia]{zhao2017pyramid}
Hengshuang Zhao, Jianping Shi, Xiaojuan Qi, Xiaogang Wang, and Jiaya Jia.
\newblock Pyramid scene parsing network.
\newblock In \emph{CVPR}, 2017.

\end{thebibliography}
\end{document}